\documentclass{article}
\usepackage{spconf,amsmath,graphicx}
\usepackage{multirow}
\usepackage{booktabs}
\usepackage{multicol}
\usepackage{url}     
\usepackage{float}
\usepackage{rotating} 

\title{One-Cycle Pruning: Pruning ConvNets with Tight Training Budget}

\name{Nathan Hubens $^{\star \dagger}$ \thanks{This research has been conducted in the context of a joint-PhD between the two institutions.} \qquad Matei Mancas $^{\star}$ \qquad  Bernard Gosselin $^{\star}$  \qquad  Marius Preda $^{\dagger}$  \qquad Titus Zaharia $^{\dagger}$ }
  
\address{$^{\star}$ ISIA Lab, University of Mons, Belgium \\
      $^{\dagger}$ Artemis, IP Paris, France}

\begin{document}
\maketitle
\begin{abstract}

Introducing sparsity in a convnet has been an efficient way to reduce its complexity while keeping its performance almost intact. Most of the time, sparsity is introduced using a three-stage pipeline: 1) training the model to convergence, 2) pruning the model, 3) fine-tuning the pruned model to recover performance. The last two steps are often performed iteratively, leading to reasonable results but also to a time-consuming process. In our work, we propose to remove the first step of the pipeline and to combine the two others in a single training-pruning cycle, allowing the model to jointly learn the optimal weights while being pruned. We do this by introducing a novel pruning schedule, named One-Cycle Pruning (OCP), which starts pruning from the beginning of the training, and until its very end. Experiments conducted on a variety of combinations between architectures (VGG-16, ResNet-18), datasets (CIFAR-10, CIFAR-100, Caltech-101), and sparsity values ($80\%$, $90\%$, $95\%$) show that not only OCP consistently outperforms common pruning schedules such as One-Shot, Iterative and Automated Gradual Pruning, but also that it drastically reduces the required training budget. Moreover, experiments following the Lottery Ticket Hypothesis show that OCP allows to find higher quality and more stable pruned networks.

\end{abstract}


\section{Introduction}
\label{sec:intro}

Deep neural networks are able to achieve state-of-the-art results in a wide variety of domains, including computer vision, natural language processing and speech recognition. But such achievements imply important increase in budget required both at training and inference time. More specifically, model size, run time memory and the number of computing operations are all constraints that make modern neural networks challenging to deploy on resource-constrained environments such as mobile phones or embedded devices. 

For those reasons, neural network compression and acceleration have become active fields of research. Several researchers have been interested in creating parameter-efficient architectures, by using low-rank approximations, parameters quantization, and neural network pruning.  

Recent studies have also exhibited a particular characteristic of neural networks, called the \textit{Lottery Ticket Hypothesis} \cite{lottery}, which suggests that, in regular architectures, there exists a sub-network that can be trained to at least the same level of performance as the original one, in a comparable training budget, as long as it starts from the same original conditions. Such a sub-network is said to have ``won'' at the initialization lottery and can be preserved, while other parameters of the original network can be removed using pruning methods. 

To prune a neural network, the most commonly used technique is the so-called Iterative Pruning \cite{BlalockOFG20}, requiring several cycles of pruning and fine-tuning which leads to a lengthy process. 
In this work, we propose to adopt a novel pruning schedule applied directly from the start of the initial training phase. This scheduling function gradually prunes the network during the training phase, thus making training and pruning of a neural network a joint process. 

Contributions of our work are summarized as following: 
\begin{itemize}
    \item We propose a novel pruning schedule with stable, thus generic parameters. 
    \item We show that the proposed pruning schedule, performed during the initial training of a neural network, drastically reduces the required training budget.
    \item We empirically show that our pruning schedule is able to find more stable lottery tickets, both for low and extreme sparsity levels, than other commonly used pruning schedules.
\end{itemize}

\section{Related Work}
Pruning techniques can differ in many aspects. The main points of differentiation are presented in this section.

\textbf{Granularity. } The granularity used for the pruning is often categorized into two groups: \textit{unstructured}, or when the pruning focuses on removing individual weights in the network and that there is no intent to keep any structure in the filters \cite{brain_damage, brain_surgeon2}. Such a pruning method leads to sparse weight matrices, requiring dedicated hardware or software to take advantage of the speed and computation gains. To overcome this limitation, \textit{structured} pruning was introduced \cite{Li, cpa, hub}, which takes care of removing complete blocks of weights such as vectors, kernels, or even convolution filters.

\textbf{Criteria. } Early works on pruning criteria make use of second-order approximation of the loss surface to remove parameters \cite{brain_damage, brain_surgeon2}. Other works have also explored the use of $l_0$ regularization \cite{l0} during the training or even the use of variational dropout \cite{var_dropout}. In addition to being more complex, those criteria often show to be less consistent across datasets and to lead to comparable or worse results than simple magnitude pruning, based on $l_1$-norm \cite{state}. Moreover, the criteria can be used to compare weights belonging to the same layer, \textit{i.e.} \textit{local} pruning, leading to layers of equivalent sparsities. The criteria can also be applied to the whole network, comparing weights from all layers, \textit{i.e.} \textit{global} pruning, and leading  to  a  network  with  layers  of  different  sparsity  levels. 

\textbf{Scheduling. } While early pruning methods cared about removing redundant weights in a single step, called one-shot pruning \cite{Li}, the most adopted technique nowadays is to perform pruning iteratively \cite{han}, starting from a pretrained network and performing several steps of pruning followed by fine-tuning. While having undeniably shown good results, such a pruning schedule is usually time consuming to obtain the final, pruned network \cite{Li}. For this reason, researchers have proposed alternatives to perform pruning during the training or even before it even started \cite{snip, to_prune}. Recently, some work has shown that the most critical phase of the training of a neural network happens during the very first iterations \cite{Frankle2020The, achille2018critical} and that applying regularization after that initial phase has little effect on the final performance of the network \cite{timematters}. One thus should apply pruning early in training to take advantage of its regularization effects but must do so very carefully to not irremediably damage the network during this brittle period \cite{stablth, lottery2}. 

For those reasons, we propose a pruning schedule that is applied gradually during the training, and designed to be gentle during the very first iterations. This pruning schedule can be used for any granularity and any selection criteria but in this work, we are focusing on unstructured pruning, \textit{i.e.} removing individual weights, and the criteria used for weights selection is their $l_1$-norms, compared locally in each layer.

\section{The Proposed Schedule}
\label{sec:sched}

\begin{figure}[h]
\begin{minipage}[b]{.49\linewidth}
  \centering
  \centerline{\includegraphics[width=4.8cm]{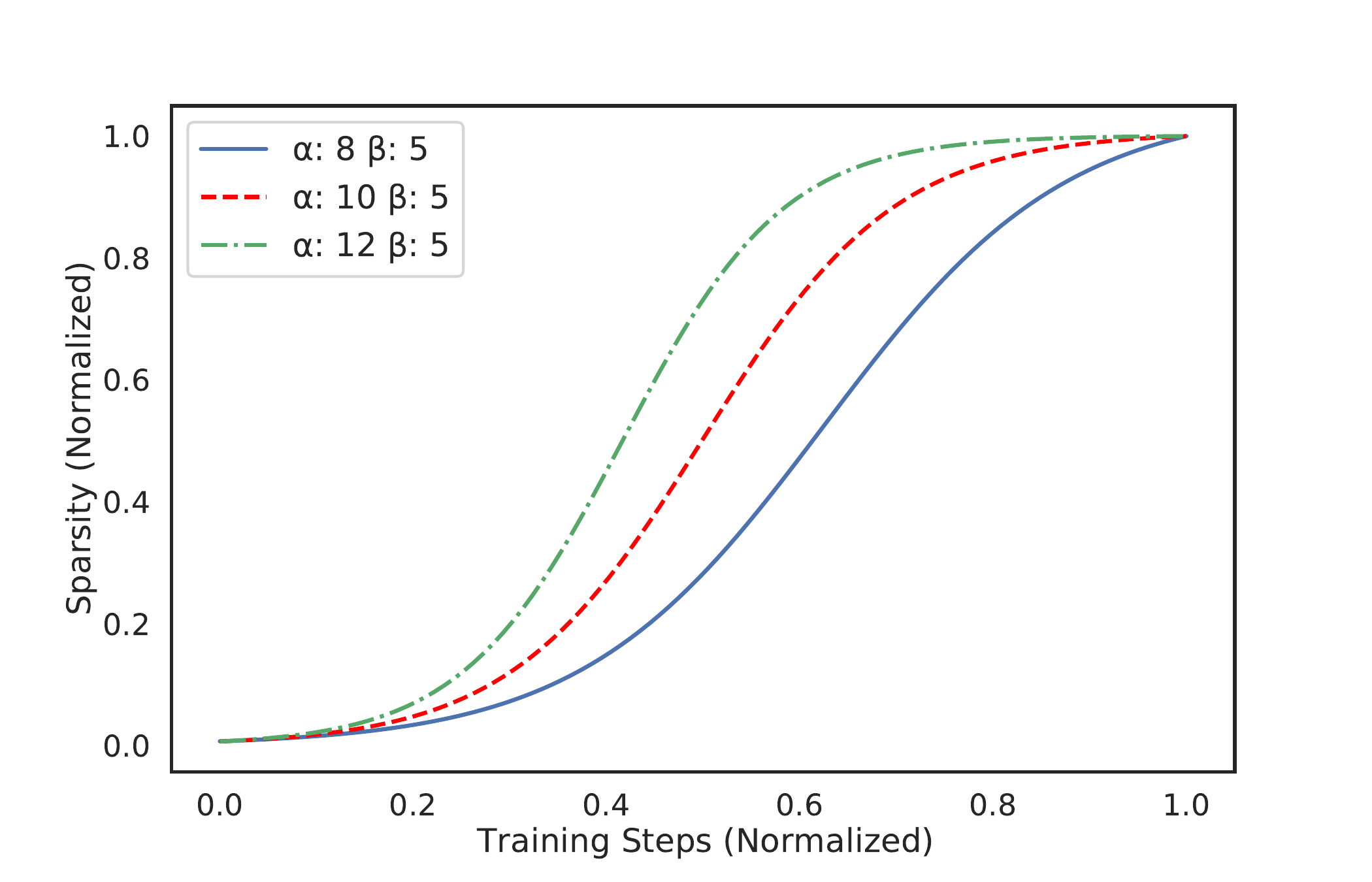}}
  {(a)}
\end{minipage}
\hfill
\begin{minipage}[b]{0.49\linewidth}
  \centering
  \centerline{\includegraphics[width=4.8cm]{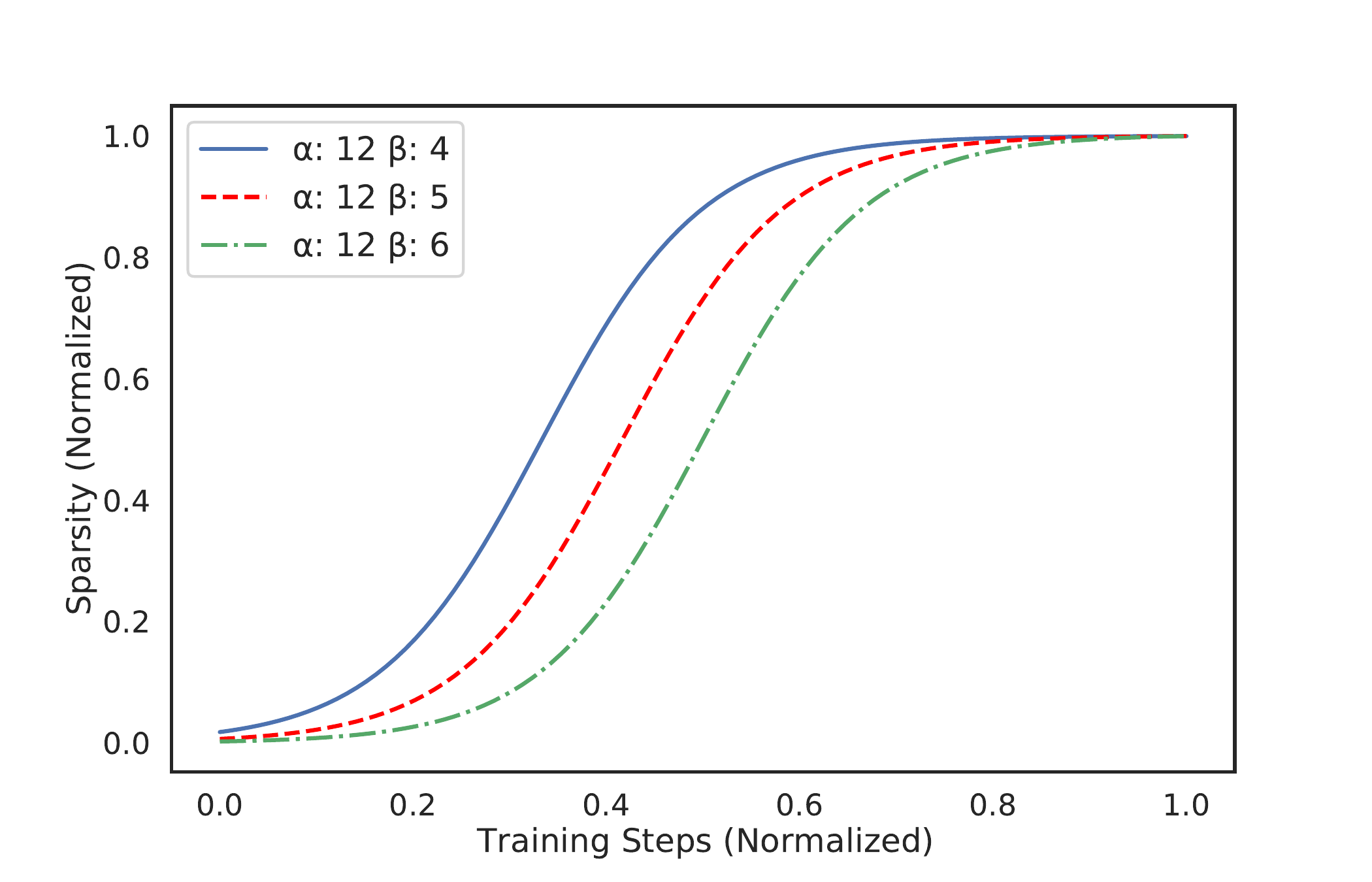}}
  {(b)}
  \end{minipage}
  \caption{Visualization of the variation of the One-Cycle Pruning schedule for different $\alpha$ and $\beta$ values.}
\label{parameters}
\end{figure}

The proposed method, One-Cycle Pruning (OCP), consists in starting from a dense network and inducing sparsity during the whole training phase. More precisely, sparsity is induced in the network according to the following schedule: 

\begin{equation}
    s_t = s_i + (s_f - s_i) \cdot \frac{1+e^{-\alpha+\beta}}{1+e^{-\alpha t + \beta}}
\label{schedule}
\end{equation}

with $s_t$, the level of sparsity at training step $t$, $s_i$ and $s_f$ respectively the initial and final level of sparsity, and $\alpha$, $\beta$ being two tuning parameters, modifying either the steepness of the scheduling (Figure \ref{parameters}a), or its horizontal offset (Figure \ref{parameters}b), to better suit the problem or architecture that is used. 

We performed a grid-search to find the best pair of $\alpha$ and $\beta$ values and found it to be $\alpha=14$ and $\beta=5$. As reported in Table \ref{parameters}, OCP is stable around its optimal parameter values. They thus can be used as default, and being adapted if needed.

\begin{table}[h]
\scriptsize
  \begin{center}
    \setlength\tabcolsep{2.5pt}
    \begin{tabular}{cc|c|c|c|c|c|l}
    \cline{3-7}
    & & \multicolumn{5}{ c| }{$\beta$} \\ \cline{3-7}
    & & 3 & 4 & 5 & 6 & 7 \\ \cline{1-7}
    \multicolumn{1}{ |c  }{\multirow{3}{*}{$\alpha$} } &
    \multicolumn{1}{ |c| }{13} & 93.10 $\pm$ 0.18 & 93.23 $\pm$ 0.12 & 93.30 $\pm$ 0.07 & 93.31 $\pm$ 0.05 & 92.97 $\pm$ 0.07 &    \\ \cline{2-7}
    \multicolumn{1}{ |c  }{}                        &
    \multicolumn{1}{ |c| }{14} & 93.13 $\pm$ 0.06 & 93.16 $\pm$ 0.07 & \textbf{93.46 $\pm$ 0.13} & 93.09 $\pm$ 0.18 & 93.25 $\pm$ 0.12 &     \\ \cline{2-7}                                     
    \multicolumn{1}{ |c  }{}                        &
    \multicolumn{1}{ |c| }{15} & 93.10 $\pm$ 0.03 & 93.21 $\pm$ 0.14 & 93.19 $\pm$ 0.19 & 93.17 $\pm$ 0.08 & 93.28 $\pm$ 0.03&      \\ \cline{1-7} \\
    \end{tabular}
  \end{center}   
      \caption{Grid search of $\alpha$ and $\beta$ for Resnet-18 trained on CIFAR-10  for 90\% sparsity. Mean and standard deviation of validation accuracy over 3 rounds are reported.}
    \label{parameters_sched}
\end{table}

The benefits of such a schedule are three-fold. First, it does not require to train the network before applying pruning, thus greatly reducing training time. Second, it prunes weights while the network is training, making the remaining weights jointly optimize for the task at hand, and the removal of other weights. Third, the design of the schedule helps the training dynamics by pruning very gently at the beginning and also helps the network to recover from pruning by slowly reducing pruning intensity towards the end.

\section{Comparison to traditional schedules}
\label{sec:comparison}

In this section, we compare our proposed schedule, the One-Cycle Pruning, to other commonly used pruning schedules.

\begin{table*}[h!]
\scriptsize
\begin{center}
\begin{tabular}{l|l||c c c c||c c c c}
\multicolumn{3}{c}{} & \multicolumn{2}{c}{\textbf{ResNet-18}} &
\multicolumn{2}{c}{}&
\multicolumn{2}{c}{\textbf{VGG-16}}\\
\cmidrule[1pt]{3-10}
\multicolumn{2}{c}{}&\multicolumn{1}{c}{One-Shot }&\multicolumn{1}{c}{Iterative }&\multicolumn{1}{c}{AGP}&\multicolumn{1}{c||}{One-Cycle}&\multicolumn{1}{c}{One-Shot }&\multicolumn{1}{c}{Iterative }&\multicolumn{1}{c}{AGP}&\multicolumn{1}{c}{One-Cycle}\\\midrule
\multicolumn{2}{l}{\textbf{CIFAR-10}} & \multicolumn{4}{c}{}               \\\midrule
\multirow{3}{4mm}{\begin{sideways}\parbox{8mm}{Sparsity}\end{sideways}}
& 80\% & 93.10 $\pm$ 0.03 & 93.13 $\pm$ 0.03  & 93.22 $\pm$ 0.22 & \textbf{93.49 $\pm$ 0.14}  & 90.25 $\pm$ 0.14 & 90.64 $\pm$ 0.19 & \textbf{90.87 $\pm$ 0.15} & 90.84 $\pm$ 0.09\\
& 90\% & 92.42 $\pm$ 0.21  & 91.72 $\pm$ 0.08 & 92.85 $\pm$ 0.09  & \textbf{93.31 $\pm$ 0.20}  & 89.82 $\pm$ 0.19 & 89.76 $\pm$ 0.18 &  90.67 $\pm$ 0.25 & \textbf{90.72 $\pm$ 0.40}\\
& 95\% & 91.58 $\pm$ 0.04 & 87.54 $\pm$ 0.39 & 92.04 $\pm$ 0.07 & \textbf{92.76 $\pm$ 0.16} & 89.73 $\pm$ 0.37 & 81.46 $\pm$ 2.87 & 90.56 $\pm$ 0.31 & \textbf{90.67 $\pm$ 0.11} \\\midrule

\multicolumn{2}{l}{\textbf{CIFAR-100}} & \multicolumn{4}{c}{}               \\\midrule
\multirow{3}{4mm}{\begin{sideways}\parbox{8mm}{Sparsity}\end{sideways}}
& 80\% & 74.21 $\pm$ 0.09  & 74.18 $\pm$ 0.29 & 74.78 $\pm$ 0.09 & \textbf{74.81 $\pm$ 0.16}  & 67.83 $\pm$ 0.19 & 67.80 $\pm$ 0.15 & 67.93 $\pm$ 0.06 & \textbf{68.34 $\pm$ 0.38} \\
& 90\% & 73.34 $\pm$ 0.23  & 71.80 $\pm$ 0.05  & 73.83 $\pm$ 0.41 &  \textbf{74.50 $\pm$ 0.24} & 67.33 $\pm$ 0.16 & 62.66 $\pm$ 1.31 &  67.88 $\pm$ 0.39 & \textbf{68.24 $\pm$ 0.45}\\
& 95\% &  71.68 $\pm$ 0.16 & 62.88 $\pm$ 0.27 & 71.92 $\pm$ 0.30 & \textbf{73.34 $\pm$ 0.21}  & 66.16 $\pm$ 0.49 & 61.95 $\pm$ 0.70 & \textbf{67.51 $\pm$ 0.19} & \textbf{67.51 $\pm$ 0.16} \\\midrule

\multicolumn{2}{l}{\textbf{Caltech-101}} & \multicolumn{4}{c}{}               \\\midrule
\multirow{3}{4mm}{\begin{sideways}\parbox{8mm}{Sparsity}\end{sideways}}
& 80\% & 80.31 $\pm$ 0.89 & 79.78 $\pm$ 0.56 & 81.93 $\pm$ 0.85& \textbf{82.31 $\pm$ 0.88}  & 77.81 $\pm$ 0.96 & 78.23 $\pm$ 0.35 & 78.45 $\pm$ 0.85 & \textbf{78.90 $\pm$ 0.88} \\
& 90\% & 79.87 $\pm$ 0.54 & 77.84 $\pm$ 0.31 & 80.89 $\pm$ 0.90 & \textbf{81.84 $\pm$ 0.16} & 78.77 $\pm$ 1.06 & 74.42 $\pm$ 2.79 & \textbf{78.57 $\pm$ 0.21} & 78.56 $\pm$ 0.31\\
& 95\% & 78.57 $\pm$ 1.02 & 73.83 $\pm$ 1.28  & 78.76 $\pm$ 1.27  & \textbf{79.81 $\pm$ 0.92} & 76.99 $\pm$ 0.78 & 42.61 $\pm$ 2.60 & 78.68 $\pm$ 0.53 & \textbf{78.99 $\pm$ 0.50} \\\midrule

\end{tabular}
\end{center}
\vspace{-3mm}
\caption{\small Results of pruning ResNet-18 and VGG-16 with 4 different schedules. Mean and standard deviation of accuracy over 3 rounds are reported. Best results are in bold.}
\label{table:results}
\end{table*}

\begin{figure}[!htbp]
 \centering
  \centerline{\includegraphics[width=7.6cm]{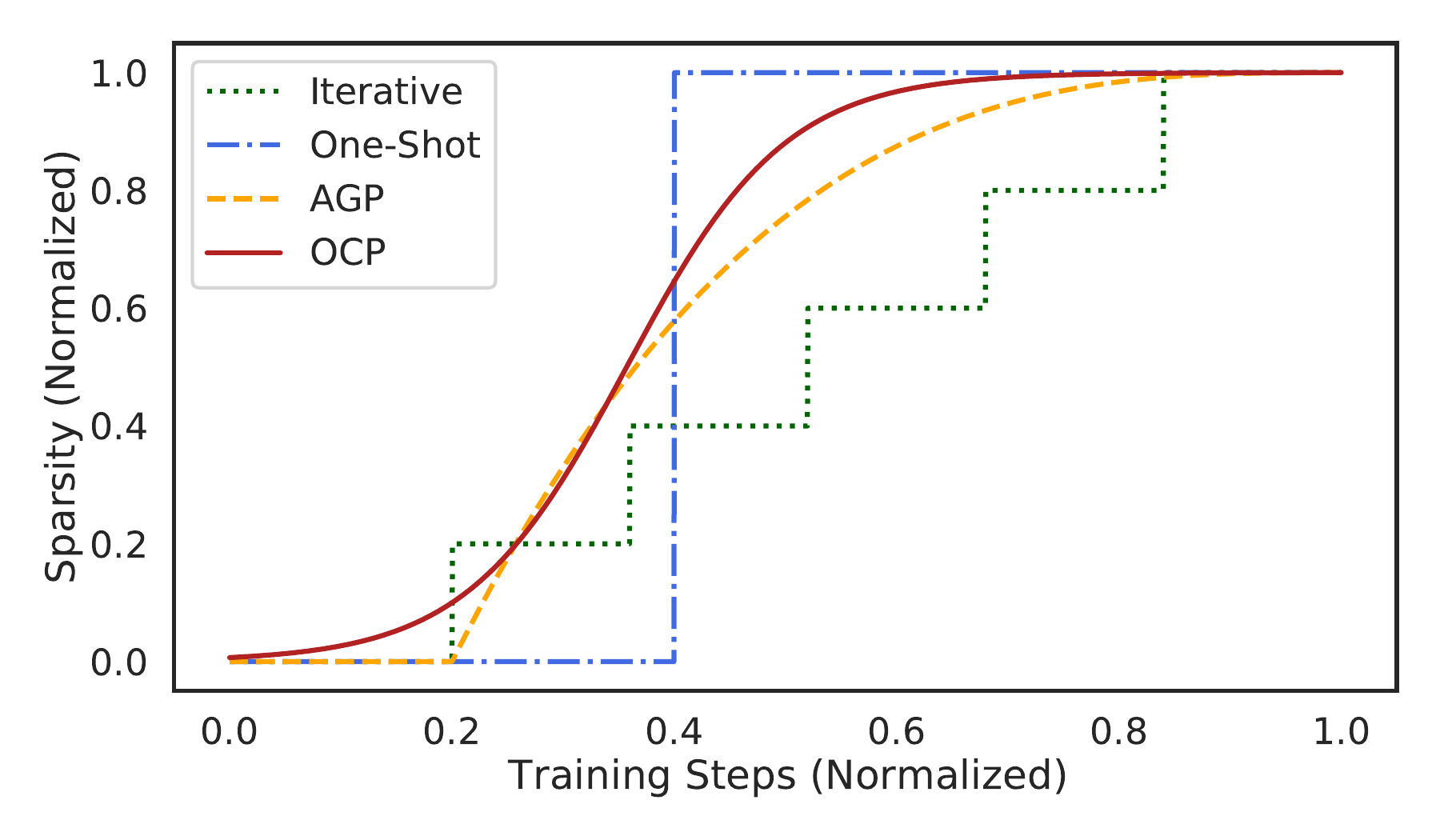}}
  \caption{Evolution of sparsity during the training of the 4 studied pruning schedules. Best results are obtained when One-Shot, Iterative and AGP start at $40\%$, $20\%$ and $20\%$ of training budget, respectively. OCP is applied right from the start.}
  \label{figure:schedules}
\end{figure}

\textbf{Pruning Methods.} We compare our pruning technique to several state-of-the-art pruning schedules: One-Shot Pruning, Iterative Pruning and Automated Gradual Pruning (AGP), under a fixed training budget. The optimal training iteration at which the pruning process starts for those schedules, \textit{i.e.} the pretraining phase, is determined by a grid search. We find the optimal starting iteration at $40\%$, $20\%$ and $20\%$ of training budget for the One-Shot Pruning, Iterative Pruning and AGP, respectively, as depicted in Figure \ref{figure:schedules}.

\textbf{Datasets and Architectures.} For our experiments, the datasets have been chosen to be various in terms of image resolution and number of classes. In particular, we evaluate our methods on the three following datasets: CIFAR-10 \cite{cifar}, CIFAR-100 \cite{cifar} and Caltech-101 \cite{caltech}. Moreover, those datasets are tested on two types of popular convolutional network architectures: VGG-16 \cite{vgg} and ResNet-18 \cite{rn18}. 

\textbf{Training Procedure.} The networks we use for our experiments are trained from a random initialization. The images are first resized to $224\times 224$ and are augmented by using horizontal flips, rotations, image warping and random cropping. We train each model using a learning rate warmup until a nominal value of $0.001$, then a gradual decay until the end of the training.

\textbf{Frameworks and Hardware.} Experiments are conducted using the PyTorch \cite{pytorch} and fastai \cite{fastai} deep learning libraries for training procedures, fasterai \cite{fasterai} library for the implementation of the pruning methods and using a 12GB Nvidia GeForce GTX 1080 Ti GPU for computation.

\textbf{Experiment A.} The first experiment is conducted with a fixed training budget of $50$ epochs. The objective is to identify the pruning schedule that is the most efficient when the training budget is tight. The results, reported in Table \ref{table:results}, show that One-Cycle Pruning usually outperforms other pruning schedules for all the different combinations of studied architectures, datasets and sparsity levels.

\textbf{Experiment B.} To better emphasize the impact of the pruning schedule on the training dynamics, we propose to fix a target accuracy and let the training budget change according to the needs of the pruning method in order to reach that accuracy level. For One-Shot and Iterative pruning, the pretraining budget is kept identical, only the fine-tuning budget is extended, \textit{i.e.} the training after the pruning has occurred. We present in Table \ref{tab:budget} the results of the training budget required to reach $90\%$, $70\%$ and $80\%$ of accuracy on CIFAR-10, CIFAR-100, Caltech-101 dataset respectively, using ResNet-18 pruned to a sparsity of $95\%$. Training budget is expressed relatively to method One-Cycle Pruning. \\

\begin{table}[h!]
  \begin{center}
    \setlength\tabcolsep{4.5pt}
    \small
    \begin{tabular}{l|cccc}
    \cmidrule[1pt]{2-5}  
      & \multicolumn{1}{c}{One-Shot} & \multicolumn{1}{c}{Iterative} & \multicolumn{1}{c}{AGP} & \multicolumn{1}{c}{One-Cycle} \\
    \hline
  {{\textbf{CIFAR-10}}} & $2.5 \times$ & $ 4\times$ & $1 \times$  & $1 \times$  \\
      
  {{\textbf{CIFAR-100}}} & $ 3.33\times$ & $ 7.5\times$ & $ 1.25 \times$  & $1 \times$  \\
      
  {{\textbf{Caltech-101}}} & $2 \times$ & $3.2 \times$ & $1.4 \times$  & $1 \times$  \\
      \bottomrule
    \end{tabular}
  \end{center}    
    \caption{Training budget required to prune ResNet-18 to $95\%$ to achieve a fixed validation accuracy of $90\%$,$70\%$ and $80\%$ on CIFAR-10, CIFAR-100 and Caltech-101 respectively}
    \label{tab:budget}
\end{table}

\textbf{Discussion.} Overall, the technique requiring the most training budget while providing the worst validation accuracy when that budget is fixed is Iterative Pruning. Several papers have also reported a similar phenomenon, indicating that Iterative Pruning required a long fine-tuning time \cite{Li, han}. On the other hand, One-Cycle Pruning seems to outperform other pruning schedules both when the training budget is fixed and when the target accuracy is fixed, indicating that our proposed schedule is able to reach higher performance faster.

\section{Experiments with Lottery Tickets}
\label{sec:LTH}

In order to investigate further in the behaviour of One-Cycle Pruning, we evaluate it using the Lottery Ticket Hypothesis.

\textbf{Finding Lottery Tickets} The Lottery Ticket Hypothesis (LTH) states that in each network, there exists a sub-network that, trained in isolation, is able to achieve comparable performance in a comparable training budget as the whole network, as long as they start from the same initialization \cite{lottery}. To discover such \textit{winning tickets}, authors propose to follow an Iterative Magnitude Pruning (IMP) method that goes as follows: an initial network, designated by a set of weight $W_0$, is trained for $t$ iterations, giving the set of weights $W_t$. Then, a binary mask $m$, whose values are either $0$ if we want to remove the corresponding weight or $1$ if we want to keep it, is applied, creating the network $W_t \odot m$, where $\odot$ denotes the element-wise product. The weights are then reinitialized to their original values $W_0 \odot m$ and the process is performed iteratively, updating $m$ until the desired sparsity is achieved. The IMP method require several rounds of one-shot pruning to unveil winning tickets. We propose to compare the quality of found tickets when they are uncovered with different pruning schedules.

\textbf{Comparison of Tickets.} We conduct the LTH experiments using ResNet-18 and CIFAR-10 with the same hyperparameters as described in Section \ref{sec:comparison}. In the original LTH experiments, IMP was applied in a single step at the end of each training round. In our experiments, we apply the studied schedules during each training round, such that our One-Shot Pruning corresponds to the original IMP experiment. We report the accuracy of the ticket of the last round of pruning in Table \ref{tab:schedules_LTH}, \textit{i.e.} $W_0 \odot m$, where $m$ is composed at $95\%$ of $0$. From that experiment, we find that untrained sub-networks found with OCP are outperforming the ones found with other schedules. This further confirms that it is possible to find already performant sub-networks directly at initialization and that we do not necessarily need to pretrain a model before starting the pruning process.

\begin{table}[h!]
  \begin{center}
    \setlength\tabcolsep{4.5pt}
    \small
    \begin{tabular}{l@{\hskip 0.2in}cc}
    \cmidrule[1pt]{1-3}  
      {\multirow{5}{*}{\textbf{Accuracy ($\%$)}}} & \multicolumn{1}{c}{\textbf{One-Shot}} & \multicolumn{1}{c}{\textbf{Iterative}}  \\
    \cmidrule[0.5pt]{2-3}
   & 72.213 $\pm$ 0.009 & 85.778 $\pm$ 0.003  \\
  \cmidrule[1pt]{2-3}
  & \multicolumn{1}{c}{\textbf{AGP}} & \multicolumn{1}{c}{\textbf{OCP}} \\
    \cmidrule[0.5pt]{2-3}  
   &  86.114 $\pm$ 0.001  & \textbf{89.239 $\pm$ 0.004}  \\
      \bottomrule
    \end{tabular}
  \end{center}    
    \caption{Validation accuracy of the reinitialized ResNet-18 sub-networks found by several pruning schedules at their last pruning round, \textit{i.e.} at a sparsity level of $95\%$. The training is performed on the CIFAR-10 dataset.}
    \label{tab:schedules_LTH}
\end{table}

\begin{figure}[!htbp]
 \centering
  \centerline{\includegraphics[width=8cm]{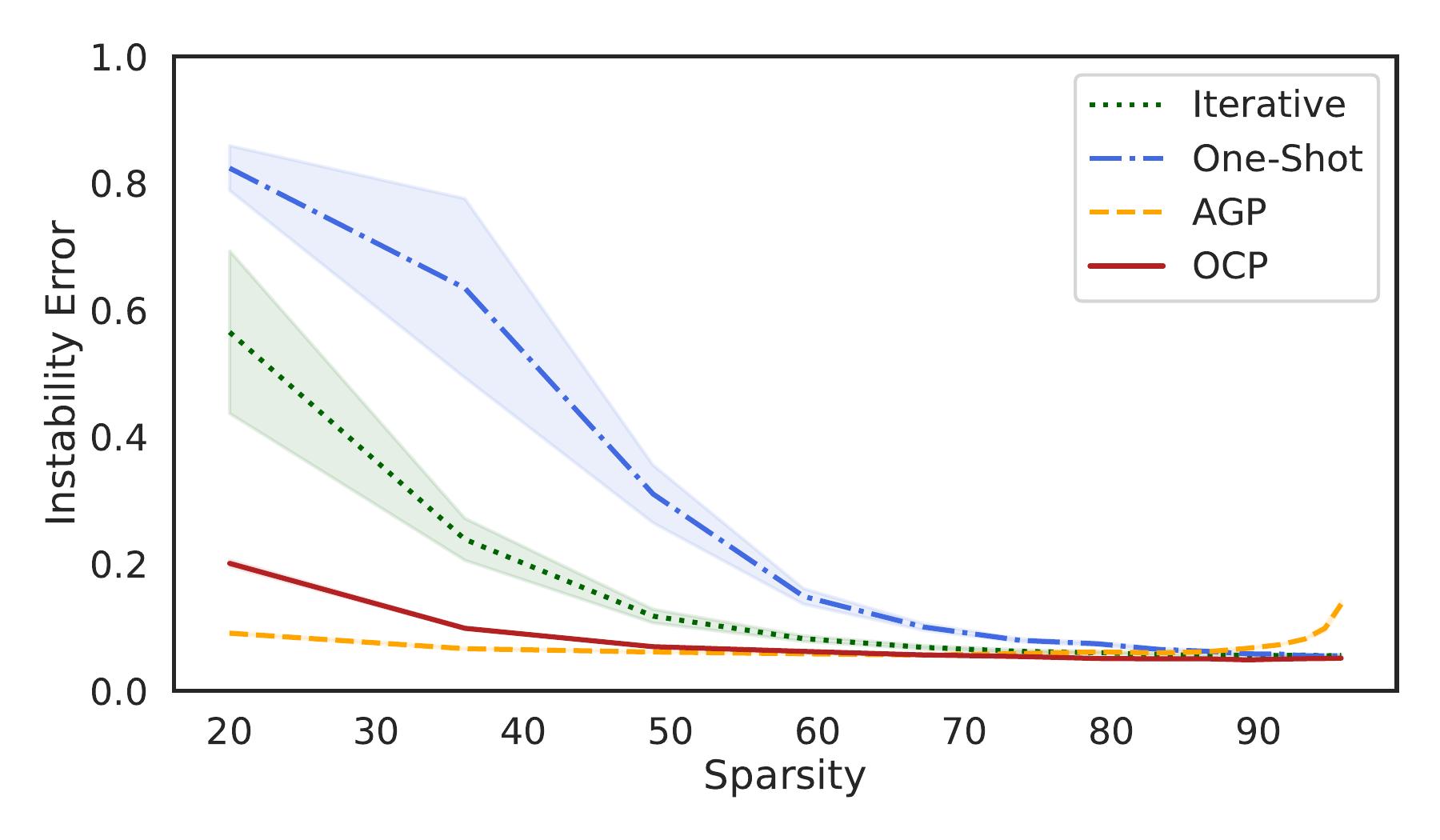}}
  \caption{Evolution of instability error after each round of the Lottery Ticket Hypothesis when using a different pruning schedule. A low instability error indicates that the found ticket $W_0 \odot m$ is stable to retraining for that particular sparsity level.}
  \label{figure:stability}
\end{figure}

\textbf{Stability of Tickets.} To further study the difference in the results, we conduct a stability analysis of the found Lottery Tickets \cite{lottery2}. This analysis consists of retraining two copies of the same ticket $W_0 \odot m$, leading to two different trained versions $W_1 \odot m$ and $W_2 \odot m$, then linearly interpolate their set of weights, setting a new network with weights $W_3 \odot m = \alpha (W_1 \odot m) + (1-\alpha)( W_2 \odot m)$ with $\alpha \in [0,1]$. The two copies will then be said linearly connected, \textit{i.e.} they converged to the same linearly connected minimum, if the validation error of $W_3 \odot m$ remains stable for all the $\alpha$ values. We report in Figure \ref{figure:stability} the evolution of instability against the level of sparsity. We denote instability error the validation error, \textit{i.e.} $1-$accuracy, with network $W_3 \odot m$, whose weights were interpolated at half-way between the two trained copies, \textit{i.e.} for $\alpha=0.5$. We can observe that OCP, as well as AGP, become stable at really low level of sparsity, \textit{i.e.} after performing only a few rounds of LTH, while One-Shot Pruning and Iterative Pruning take more rounds before getting stable. However, AGP seems to be the only one to show signs of increasing instability for high sparsity values, \textit{i.e.} for late LTH pruning rounds.

\section{Conclusion}
\label{sec:ccl}

 In this work, we proposed One-Cycle Pruning, a novel pruning schedule that allows a network to be pruned in a single training-pruning phase, removing the needs of an initial pretraining phase but also of a complex and time-consuming fine-tuning phase. When compared to common pruning schedules, One-Cycle Pruning provides comparable or better results with significantly less computation required. We also studied the quality of Lottery Tickets found with One-Cycle Pruning. We showed that those tickets were of higher quality and were consistently stable both at low and high level of sparsities, further validating the benefits of the proposed pruning schedule.
 




\bibliographystyle{IEEEbib}
\bibliography{biblio}

\appendix
\end{document}